\documentclass[letterpaper]{article} 
\usepackage{aaai2026}  
\usepackage{times}  
\usepackage{helvet}  
\usepackage{courier}  
\usepackage[hyphens]{url}  
\usepackage{graphicx} 
\usepackage{natbib}  
\usepackage{caption} 
\usepackage{algorithm}
\usepackage{algorithmic}

\usepackage{amsmath}
\usepackage{booktabs}
\usepackage{natbib}

\usepackage{color}
\usepackage{listings}
\usepackage{caption}

\usepackage{newfloat}
\usepackage{listings}
\DeclareCaptionStyle{ruled}{labelfont=normalfont,labelsep=colon,strut=off} 
\floatstyle{ruled}
\newfloat{listing}{tb}{lst}{}
\floatname{listing}{Listing}
\title{Few-Shot and Training-Free Review Generation via Conversational Prompting}
\author{
    Genki Kusano
}
\affiliations{
    NEC Corporation\\
    g-kusano@nec.com
}

\usepackage{bibentry}

\begin{document}

\maketitle

\begin{abstract}
Personalized review generation helps businesses understand user preferences, yet most existing approaches assume extensive review histories of the target user or require additional model training. Real-world applications often face few‑shot and training‑free situations, where only a few user reviews are available and fine‑tuning is infeasible. It is well known that large language models (LLMs) can address such low‑resource settings, but their effectiveness depends on prompt engineering. In this paper, we propose Conversational Prompting, a lightweight method that reformulates user reviews as multi‑turn conversations. Its simple variant, Simple Conversational Prompting (SCP), relies solely on the user's own reviews, while the contrastive variant, Contrastive Conversational Prompting (CCP), inserts reviews from other users or LLMs as incorrect replies and then asks the model to correct them, encouraging the model to produce text in the user's style. Experiments on eight product domains and five LLMs showed that the conventional non-conversational prompt often produced reviews similar to those written by random users, based on text-based metrics such as ROUGE-L and BERTScore, and application-oriented tasks like user identity matching and sentiment analysis. In contrast, both SCP and CCP produced reviews much closer to those of the target user, even when each user had only two reviews. CCP brings further improvements when high‑quality negative examples are available, whereas SCP remains competitive when such data cannot be collected. These results suggest that conversational prompting offers a practical solution for review generation under few‑shot and training‑free constraints.
\end{abstract}



\section{Introduction}

Product reviews play a key role in understanding user needs and supporting business decisions in e-commerce, marketing, and product design. Companies have traditionally gathered feedback through surveys or interviews, but these methods are both costly and time-consuming. To reduce this burden, researchers have explored personalized review generation~\citep{DBLP:conf/acl/prg_NiM18, DBLP:conf/aaai/rag_review_Xie0MM23}.

Conventional methods train models to generate fluent reviews by using a large amount of review data. However, as mentioned earlier, collecting such review data is often difficult and costly. Large language models (LLMs) can now create product reviews in a training‑free manner from user reviews~\citep{DBLP:journals/corr/review_llm}. Role-playing techniques~\citep{DBLP:journals/corr/persona_survey_tmlr} also show that LLMs can write text in a personal style, yet they rely on rich profile information such as long review histories~\citep{DBLP:conf/emnlp/characterllm_ShaoLDQ23}. In contrast, we focus on consumers who have only a few past reviews and this sparse condition is common in real marketing scenarios.

\begin{figure}[t]
\centering
\includegraphics[width=85mm]{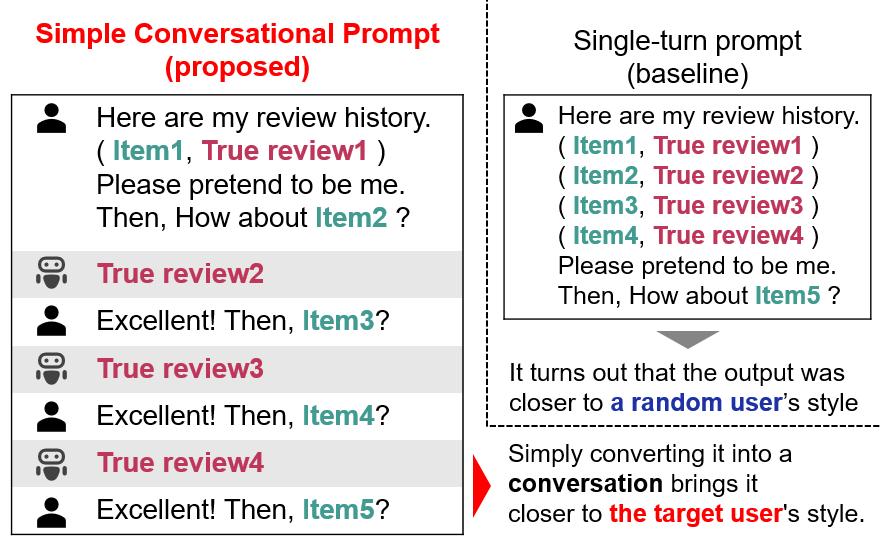}
\caption{The conventional method (right) often generates outputs similar to random reviews. The proposed method (left) converts the review history into a conversation, which helps the LLM reflect the target user's writing style. Inserting incorrect reviews can further improve output quality.}
\label{fig:intro}
\end{figure}

Under a training-free setting with limited user reviews, we propose {\em Conversational Prompting}. The approach converts a user's past reviews into a multi-turn conversational format. Each item is given as an input message, and the corresponding past review is used as the response from the LLM (Figure~\ref{fig:intro}). Through this structure, the LLM can imitate the writing style of the target user more effectively. If reviews written by other users are available, they can be added to the prompt as incorrect responses. After each incorrect response, a message states that it does not match the user style, which helps the LLM focus on the user's preferences and writing style by comparing correct and incorrect responses.

In this paper, inspired by contrastive learning~\citep{DBLP:conf/cvpr/contrastive_HadsellCL06}, we refer to the method that adds other-user reviews as negative examples in the conversation as \textbf{C}ontrastive \textbf{C}onversational \textbf{P}rompting (\textbf{CCP}). The version without negative examples is called \textbf{S}imple \textbf{C}onversational \textbf{P}rompting (\textbf{SCP}). There is a clear difference between the two methods: CCP requires external reviews, whereas SCP does not require them.

We evaluated the proposed methods, SCP and CCP, on eight real-world datasets. We varied user history length from 2 to 10 reviews and compared five LLMs to test sensitivity to data size and model choice. Fewer than five reviews correspond to a cold-start scenario typical of long-tail users, where producing high-quality reviews under limited data is important for deeper user understanding. For quality assessment, we used ROUGE-L~\citep{lin-2004-rouge} to assess word overlap and BERTScore~\citep{DBLP:conf/iclr/bertscore_ZhangKWWA20} to capture semantic closeness. We also introduced two application-oriented downstream tasks: user identification and sentiment analysis. These tasks were useful not only for review generation but also for other user-centric evaluations. Results showed that the non-conversational prompt often produced reviews similar to those from random users, while both SCP and CCP generated reviews closer to the target user. CCP slightly outperformed SCP in some settings, but the difference was small, and there is a trade-off between accuracy and the cost of collecting negative examples. These results suggest that both methods are reasonable choices, and the choice should depend on the specific application context.

\section{Related Work}
Traditional approaches to personalized review generation have used supervised models that learn user-specific expressions from large review datasets. Early studies introduced probabilistic generation using user and item IDs~\citep{DBLP:conf/ijcnlp/lstm_NiLVM17} and encoder-decoder models~\citep{DBLP:conf/acl/prg_NiM18}. Other work explored style control with GANs~\citep{DBLP:conf/emnlp/prggan_LiT19} and aspect-aware generation with coarse-to-fine strategies~\citep{DBLP:conf/acl/coarse_LiZWS19}. Some models jointly learned recommendation and review generation within a dual-learning framework~\citep{DBLP:conf/www/dual_Xrec_SunWZFHW20}.

In the context of review generation, explainable recommendation has also been studied, where the goal is to generate reasons for recommending items. Related approaches include multi-task learning for rating prediction and explanation generation~\citep{DBLP:conf/recsys/whyilike_LuDS18}, retrieval to identify aspects~\citep{DBLP:conf/acl/exp_rec_ChengWLZ0LL23}, retrieval to reduce hallucination~\citep{DBLP:conf/aaai/rag_review_Xie0MM23}, and the use of graph neural networks and LLMs~\citep{DBLP:conf/emnlp/xrec_MaR024}.

LLMs have been used for unsupervised review generation and explainable recommendation, which are part of the broader trend of personalized content generation~\cite{DBLP:conf/acl/personalized_XuZSHWFZ0C25}. Some work has compared the outputs of LLMs with traditional explainable recommendation methods~\citep{DBLP:journals/corr/gptbaseline_abs-2309-08817, DBLP:journals/corr/23_is_chatgpt}. Review-LLM~\citep{DBLP:journals/corr/review_llm} is a method that generates reviews using prompts without training. Other studies have proposed role-playing methods that imitate the behavior of a target user based on long review histories or personality traits~\citep{DBLP:journals/corr/persona_survey_tmlr, DBLP:conf/emnlp/characterllm_ShaoLDQ23, DBLP:conf/acl/role_llm_WangPQLZWGGN00024, DBLP:conf/emnlp/harry_potter_ChenWJ0LCWL23}. One approach also suggests learning personality features from psychological measures and distilling them from large role-playing agents~\citep{DBLP:conf/emnlp/role_play_mind_RanWXYLXY24}. These methods often require long user review histories, so they may not perform well in few-shot settings. Therefore, this study addresses that limitation.

\section{Product Review Generation}

\subsection{Problem Setting}
For a user $u$ and item $i_k$, let $r^u_k$ be the review written by $u$ for $i_k$, and let $\{(i_k, r^u_k)\}_{k=1}^{n}$ denote the chronological review history of $u$, where $i_n$ is the most recent. Our goal is to generate a review $\hat{r}^u_{n+1}$ for the item $i_{n+1}$ that $u$ has not reviewed yet. Ideally, $\hat{r}^u_{n+1}$ should closely resemble the actual review that $u$ would write for the item.

\subsection{Multi-turn conversation}
We revisit the construction for prompts of LLMs as multi-turn conversations. A message is defined as 
\[
M(role, ~ text) := \{\text{`role'}: role,\ \text{`content'}: text\},
\]
where $role \in \{\texttt{user}, \texttt{assistant}\}$ indicates the speaker\footnote{LLMs typically use two main roles: \texttt{user} and \texttt{assistant}, as seen in OpenAI and Anthropic models. Other LLMs, such as the \texttt{llama} series, do not follow the same role names but support similar role-based prompting. For consistency, we use \texttt{user} and \texttt{assistant} to describe all prompts in this paper.} and $text$ is the corresponding content. A sequence of such messages forms a prompt called a multi-turn conversation\footnote{A single-turn prompt, i.e., $[M(\texttt{user}, ~ text)]$, corresponds to the conventional definition of a prompt.}.

\begin{figure}[t]
\centering
\includegraphics[width=83mm]{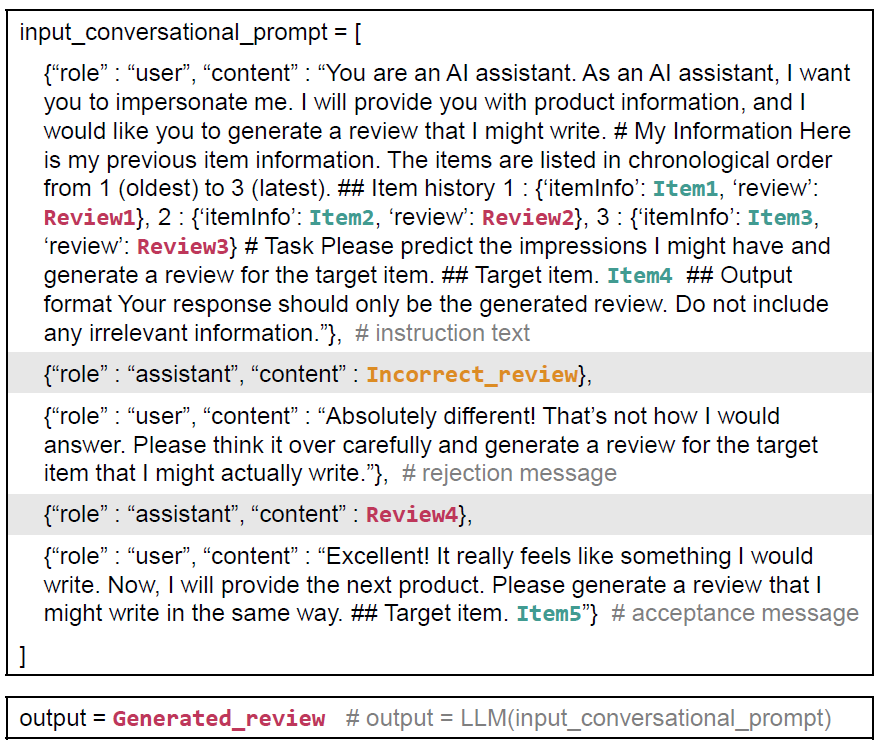}
\caption{An example of a conversational prompt using CCP, starting with the first instruction text $T_{3}$, is followed by an incorrect review $r'_4$, a rejection message, the true review $r^u_4$, and an acceptance message that requests a review for $i_5$.}
\label{fig:prompts}
\end{figure}

\subsection{SCP: Simple Conversational Prompting}
We define the first instruction text $T_{n-\ell}$ ($0 \leq \ell < n$), which asks the LLM to generate a review for item $i_{n-\ell+1}$ based on the past review history $\{(i_k, r^u_k)\}_{k=1}^{n-\ell}$, as shown in Figure~\ref{fig:prompts}. In conventional in-context learning, all $n$ past reviews are concatenated into a single user message, $[M(\texttt{user}, T_n)]$. In contrast, our proposed method, \textit{Simple Conversational Prompting} (SCP), divides the same information into $\ell$ rounds of positive feedback interactions.

In SCP, the first instruction text $M(\texttt{user}, T_{k-1})$ starts the conversation. The \texttt{assistant} role then provides the actual review $r^u_k$, simulating the target user. The \texttt{user} role replies with an acceptance message and asks for a review of the next item $i_{k+1}$ (see Figure~\ref{fig:prompts} for the detailed wording). This interaction is repeated until the final \texttt{user} message asks for a review of the target item $i_{n+1}$. Because SCP only converts the review history into a conversational format, it does not require additional data or inference cost.

\subsection{CCP: Contrastive Conversational Prompting}
SCP was designed as a fully unsupervised method. However, if another review for the item $i_k$ is available, it can be added as follows. When the \texttt{user} requests a review for item $i_k$, the \texttt{assistant} intentionally responds with a review $r'_k$ written by another user ($r'_k \neq r^u_k$) instead of the true review. The \texttt{user} then rejects this incorrect response and asks the \texttt{assistant} to regenerate the review for the same item (see Figure~\ref{fig:prompts} for the detailed wording). After the rejection, the interaction proceeds in the same manner as in SCP: the \texttt{assistant} outputs the correct review $r^u_k$, and the \texttt{user} accepts it and moves on to the next item $i_{k+1}$.

This conversational structure with negative feedback is intended to encourage the LLM to distinguish between the target user's style and other users' styles, motivated by the concept of contrastive learning~\citep{DBLP:conf/cvpr/contrastive_HadsellCL06}. We call this method \textit{Contrastive Conversational Prompting} (CCP).

\section{Experimental Setup}

\subsection{Dataset}
We used eight categories from the Amazon Reviews dataset~\citep{DBLP:journals/corr/amazon23_abs-2403-03952}\footnote{We refer to Movies\_and\_TV as Movies, CDs\_and\_Vinyl as Music, Kindle\_Store as Kindle, Grocery\_and\_Gourmet\_Food as Groceries, Toys\_and\_Games as Games, respectively.} and selected users with at least six reviews and items with at least five reviews by other users. Reviews were limited to 20–300 tokens\footnote{Short reviews such as ``Good!'' are trivial to predict, while very long ones often include HTML tags, making them unsuitable.}. We then randomly sampled 200 users per category, resulting in 1,600 users across the eight categories. For each user, the most recent review was reserved for evaluation, and the other five were used as review history to form the multi-turn conversation, i.e., $n=5$.

To evaluate the quality of generated reviews, we used ROUGE-L~\citep{lin-2004-rouge} for word overlap and BERTScore~\citep{DBLP:conf/iclr/bertscore_ZhangKWWA20} for semantic similarity. Table~\ref{table:stat} shows the average ROUGE-L and BERTScore for reviews written by other users about the same items. For each item, we selected three reference reviews: the highest-scoring, a random one, and the lowest-scoring, each compared with the target user's review. The table also lists the median and mean number of reviewers per item.

Each score can be interpreted as follows. A generated review with a ROUGE score around 0.12 shows similarity comparable to a randomly selected review from approximately 60 users (median). When the ROUGE score exceeds 0.22, it matches the most aligned reviews by other users. For BERTScore, the average for random reviews is about 0.84, while the most similar ones reach 0.87. These values serve as reference points for interpreting absolute scores.

\begin{table*}[t]
\caption{Dataset statistics. The column names ``R'' and ``B'' refer to ROUGE-L and BERTScore, respectively.}
\label{table:stat}
\centering
\scalebox{0.70}{
\begin{tabular}{c|cc|cc|cc|cc|cc|cc|cc|cc||cc}
\toprule
 & \multicolumn{2}{|c}{Movies} & \multicolumn{2}{c}{Music} & \multicolumn{2}{c}{Books} & \multicolumn{2}{c}{Kindle} & \multicolumn{2}{c}{Groceries} & \multicolumn{2}{c}{Games} & \multicolumn{2}{c}{Sports} & \multicolumn{2}{c}{Electronics} & \multicolumn{2}{||c}{All} \\ \midrule
 & R & B & R & B & R & B & R & B & R & B & R & B & R & B & R & B & R & B  \\
\midrule
Max & $0.216$ & $0.868$ & $0.215$ & $0.871$ & $0.235$ & $0.876$ & $0.233$ & $0.876$ & $0.234$ & $0.874$ & $0.234$ & $0.877$ & $0.219$ & $0.871$ & $0.226$ & $0.872$ & $0.226$ & $0.873$ \\
Mean & $0.117$ & $0.841$ & $0.119$ & $0.845$ & $0.127$ & $0.849$ & $0.125$ & $0.848$ & $0.118$ & $0.844$ & $0.125$ & $0.849$ & $0.119$ & $0.846$ & $0.117$ & $0.842$ & $0.121$ & $0.845$ \\
Min & $0.039$ & $0.805$ & $0.040$ & $0.810$ & $0.046$ & $0.816$ & $0.046$ & $0.815$ & $0.026$ & $0.788$ & $0.032$ & $0.804$ & $0.037$ & $0.803$ & $0.023$ & $0.783$ & $0.036$ & $0.803$ \\ \midrule
Median (Users) & \multicolumn{2}{|c}{$39$} & \multicolumn{2}{|c}{$31$} & \multicolumn{2}{|c}{$54$} & \multicolumn{2}{|c}{$60$} & \multicolumn{2}{|c}{$124$} & \multicolumn{2}{|c}{$62$} & \multicolumn{2}{|c}{$63$} & \multicolumn{2}{|c}{$196$} & \multicolumn{2}{||c}{$60$} \\
Mean (Users) & \multicolumn{2}{|c}{$76.9 \pm 106.6$} & \multicolumn{2}{|c}{$58.9 \pm 81.1$} & \multicolumn{2}{|c}{$141.9 \pm 244.5$} & \multicolumn{2}{|c}{$135.8 \pm 195.3$} & \multicolumn{2}{|c}{$552.4 \pm 1572.8$} & \multicolumn{2}{|c}{$225.6 \pm 710.9$} & \multicolumn{2}{|c}{$307.6 \pm 703.9$} & \multicolumn{2}{|c}{$785.2 \pm 1475.3$} & \multicolumn{2}{||c}{$285.5 \pm 880.7$} \\
\bottomrule
\end{tabular}
}
\end{table*}

\subsection{Comparison Methods}

We use the prompt from Review-LLM~\citep{DBLP:journals/corr/review_llm} as the \textbf{Baseline}, which concatenates all $n$ past reviews into a single input $[M(\texttt{user}, T_{n})]$. Since we focus on a training-free setting with only one user's history, we exclude supervised methods, including the fine-tuned Review-LLM.

In the experiments, we examine how the type of negative examples affects the performance of CCP. One approach uses reviews with the highest ROUGE or BERTScore as negative examples, denoted as \textbf{CCP(R)} and \textbf{CCP(B)}, respectively. To evaluate the review quality, we also use the lowest-scoring ones, denoted as \textbf{CCP(R)$^{-}$} and \textbf{CCP(B)$^{-}$}.

Another type of negative example can be generated by the LLM itself, which we call \textbf{CCP(G)}. We first set the number of turns $0 < \ell < n$. In the initial turn, a negative review $r'_k$ ($k = n - \ell + 1$) is generated using $[M(\texttt{user}, T_{k-1})]$. In subsequent turns, we create conversations with the previous history, the negative review $r'_k$, its rejection feedback, the correct review $r^u_k$, an acceptance message, and a request for the next item $i_{k+1}$. \textbf{CCP(G)} does not require reviews from real users but increases inference cost.

As a post-processing step, \textbf{Self-Refine}~\citep{DBLP:conf/nips/MadaanTGHGW0DPY23_selfrefine}, which adds feedback to a generated output for refinement, is included as a comparison method\footnote{The specific prompt wording is provided in the appendix. In this paper, we report only the results of applying \textbf{Self-Refine} to the \textbf{Baseline}. Applying \textbf{Self-Refine} to \textbf{SCP} or \textbf{CCP} did not improve accuracy, so those results are omitted and shown in the appendix.}.

\section{Automatic Evaluation of Review Quality}
This section examines the effectiveness of the proposed prompting methods (SCP and CCP) for review generation, aiming to address the following research questions:

\begin{description}
    \item[RQ1]: Do conversational prompts improve the quality of generated reviews compared to non-conversational ones?
    \item[RQ2]: How do the number of conversational turns ($\ell$) and the amount of feedback affect generation performance?
    \item[RQ3]: How does the number of input reviews ($n$) in the user review history influence generation performance?
    \item[RQ4]: How robust are the effects across different LLMs?
\end{description}

Each experiment was run once due to the budget constraint of API inference costs. To ensure statistical reliability, we evaluated statistical significance with the Wilcoxon test and reported confidence intervals based on the $t$-distribution or the bootstrap method.

\subsection{RQ1: Effectiveness of Conversational Prompting}

\begin{table*}[t]
\caption{ROUGE-L and BERTScore, where bold and underlined values represent best and second-best methods in each column. The column names ``R'' and ``B'' refer to ROUGE-L and BERTScore, respectively. Prompts significantly better than \textbf{SCP} are marked with $*$, while those not significantly better than \textbf{Baseline} are marked with $\diamond$ (one-sided Wilcoxon test, $p < 0.01$).}
\label{table:score_dataset}
\centering
\scalebox{0.65}{
\begin{tabular}{c|cc|cc|cc|cc|cc|cc|cc|cc||cc}
\toprule
 & \multicolumn{2}{|c}{Movies} & \multicolumn{2}{c}{Music} & \multicolumn{2}{c}{Books} & \multicolumn{2}{c}{Kindle} & \multicolumn{2}{c}{Groceries} & \multicolumn{2}{c}{Games} & \multicolumn{2}{c}{Sports} & \multicolumn{2}{c}{Electronics} & \multicolumn{2}{||c}{All} \\ \midrule
 & R & B & R & B & R & B & R & B & R & B & R & B & R & B & R & B & R & B\\
\midrule
Baseline & $0.126$ & $0.844$ & $0.129$ & $0.847$ & $0.148$ & $0.852$ & $0.137$ & $0.852$ & $0.132$ & $0.846$ & $0.137$ & $0.850$ & $0.137$ & $0.848$ & $0.123$ & $0.843$ & $0.133$ & $0.848$ \\
Self-Refine & $0.141$ & $0.845^{\diamond}$ & $0.144$ & $0.849^{\diamond}$ & $0.157^{\diamond}$ & $0.854^{\diamond}$ & $0.134^{\diamond}$ & $0.850^{\diamond}$ & $0.127^{\diamond}$ & $0.845^{\diamond}$ & $0.137^{\diamond}$ & $0.851^{\diamond}$ & $0.130^{\diamond}$ & $0.846^{\diamond}$ & $0.123^{\diamond}$ & $0.842^{\diamond}$ & $0.137^{\diamond}$ & $0.848^{\diamond}$  \\ \midrule
SCP & $0.153$ & $0.854$ & $0.165$ & $0.859$ & $0.183$ & $0.864$ & $0.154$ & $0.860$ & $0.143$ & $0.854$ & $0.152$ & $0.860$ & $0.149$ & $0.857$ & $0.134$ & $0.850$ & $0.154$ & $0.857$ \\
CCP(B) & $\underline{0.171}^{*}$ & $\textbf{0.860}^{*}$ & $\underline{0.177}$ & $\underline{0.864}^{*}$ & $0.186$ & $\underline{0.866}^{*}$ & $\underline{0.159}$ & $\textbf{0.863}^{*}$ & $0.148$ & $\textbf{0.857}^{*}$ & $\textbf{0.157}$ & $\textbf{0.863}^{*}$ & $\underline{0.152}$ & $\textbf{0.858}^{*}$ & $0.136$ & $\textbf{0.853}^{*}$ & $\underline{0.161}^{*}$ & $\textbf{0.861}^{*}$ \\
CCP(R) & $0.167^{*}$ & $0.857^{*}$ & $0.175$ & $0.863^{*}$ & $\underline{0.187}$ & $0.866^{*}$ & $\textbf{0.161}$ & $\underline{0.863}^{*}$ & $\underline{0.149}$ & $\underline{0.857}^{*}$ & $\underline{0.155}$ & $0.861^{*}$ & $0.150$ & $\underline{0.857}$ & $\textbf{0.138}$ & $\underline{0.852}^{*}$ & $0.160^{*}$ & $0.860^{*}$ \\
CCP(G) & $\textbf{0.175}^{*}$ & $\underline{0.858}^{*}$ & $\textbf{0.187}^{*}$ & $\textbf{0.865}^{*}$ & $\textbf{0.191}^{*}$ & $\textbf{0.867}^{*}$ & $0.159$ & $0.863^{*}$ & $\textbf{0.150}$ & $0.856$ & $0.155$ & $\underline{0.861}$ & $\textbf{0.158}$ & $0.857$ & $\underline{0.137}$ & $0.850$ & $\textbf{0.164}^{*}$ & $\underline{0.860}^{*}$ \\ \midrule
CCP(B)$^{-}$ & $0.152$ & $0.854$ & $0.172$ & $0.862$ & $0.179$ & $0.863$ & $0.155$ & $0.861$ & $0.141$ & $0.854$ & $0.153$ & $0.861$ & $0.149$ & $0.857$ & $0.135$ & $0.851^{*}$ & $0.154$ & $0.858^{*}$ \\
CCP(R)$^{-}$  & $0.163$ & $0.857^{*}$ & $0.173$ & $0.862$ & $0.186$ & $0.866^{*}$ & $0.155$ & $0.862^{*}$ & $0.145$ & $0.856^{*}$ & $0.154$ & $0.861$ & $0.150$ & $0.857$ & $0.133$ & $0.851^{*}$ & $0.157$ & $0.859^{*}$ \\
\bottomrule
\end{tabular}
}
\end{table*}

In the following section, we use \texttt{gpt-4.1-mini}\footnote{The temperature is fixed at 0.1. Additional implementation details, including version information and per-token inference cost, are provided in the appendix.} as the LLM. Table~\ref{table:score_dataset} presents the ROUGE and BERTScore results obtained under the setting where feedback is provided for all items in user review histories except the first one, i.e., $\ell = 4$. The main findings are summarized as follows:

(1) The \textbf{Baseline} achieved ROUGE scores of 0.12–0.15 and BERTScores of 0.843–0.852. As shown in Table~\ref{table:stat}, these values are comparable to reviews randomly written by other users for the same items. \textbf{Self-Refine} was rarely better than \textbf{Baseline} and did not improve review generation. Therefore, non-conversational prompts reflected general opinions about an item but failed to capture the target user's writing style.

(2) As there were no $\diamond$ marks in Table~\ref{table:score_dataset}, \textbf{SCP} and \textbf{CCP} significantly outperformed the \textbf{Baseline}. Across all datasets, both methods achieved ROUGE scores above 0.15 and BERTScores above 0.855, indicating that the generated reviews were closer to the target user's style than random reviews. In some datasets, \textbf{CCP($\cdot$)} ($\cdot \in \{\textbf{B}, \textbf{R}, \textbf{G}\}$) reached ROUGE scores over 0.18 and BERTScores over 0.86. Compared to the results in (1), these findings support the effectiveness of conversational prompting for review generation.

(3) Among the \textbf{CCP} variants that use other users' reviews as negative examples, the quality of these negatives had a strong effect on performance. High-quality \textbf{CCP($\cdot$)} ($\cdot \in \{\textbf{B}, \textbf{R}\}$) performed better than \textbf{SCP} in many cases. In contrast, low-quality \textbf{CCP($\cdot$)}$^{-}$ led to only small improvements. The results support ``hard negatives''~\citep{DBLP:conf/iclr/hard_negative_RobinsonCSJ21} that are close to the decision boundary can help the model learn better, while examples that are very different are less useful. In addition, \textbf{CCP(B)} often performed better than \textbf{CCP(R)}. This suggests that BERTScore may be a better way to select negative examples than ROUGE.

(4) \textbf{CCP(G)} also statistically outperformed \textbf{SCP} in many cases. Its performance was close to \textbf{CCP(B)}, suggesting that the negative examples generated by the LLM can be as effective as human-written hard negatives.

\subsubsection{Examples of Generated Reviews}

Table~\ref{table:review_example} presents the target user's actual review in the Music category (True), the reviews generated by each prompting method, and the review history used for generation. The original review included the phrases ``unique sound'' and ``a very good album'', which were reflected in the outputs of \textbf{SCP} and \textbf{CCP} through expressions such as ``have such a unique sound'' and ``This collection is a great way''. Similar phrasing appeared in the review history, indicating that the conversational prompts captured the user's writing style. In contrast, the \textbf{Baseline} focused on detailed product descriptions and did not retain the user's tone and expressions. These observations suggest that the conversational prompts reproduced the user's style more effectively than the non-conversational prompt.

\begin{table}[t]
\centering
\caption{Actual and generated reviews along with the input review history. The numerical value at the end of each generated review indicates BERTScore.}
\label{table:review_example}
\scalebox{0.76}{
\begin{tabular}{p{1.24\linewidth}}
\toprule

\textbf{True}: The Moody Blues were such \textbf{a unique sound} when they came out.  I enjoy listening to their music and this is \underline{a very good album}.\\
\midrule

\textbf{SCP}: The Moody Blues have such \textbf{a unique sound} that really stands out in rock music.  \underline{This collection is a great way} to enjoy their best songs all in one place.  The remastering sounds excellent and the booklet is a nice bonus. (0.913) \\
\midrule

\textbf{CCP(B)}: The Moody Blues have such \textbf{a unique sound} that I always enjoy. \underline{This collection has all the classics I love}, and the remastering really brings out the best in their music. A great compilation for both longtime fans and newcomers. (0.913) \\
\midrule

\textbf{CCP(G)}: The Moody Blues have such \textbf{a unique sound} that I always enjoy. \underline{This collection has a great mix} of their best songs, and the remastering really brings out the details. A nice set for both longtime fans and new listeners. (0.911) \\
\midrule

\textbf{Baseline}: The Very Best Of The Moody Blues is \underline{a fantastic collection} that really showcases the band's incredible range and timeless sound. I love how it spans their entire career, from the early days with "Go Now" to later hits like "I Know You're Out There Somewhere." The remastered tracks sound crisp and vibrant, bringing new life to classics like "Nights in White Satin" and "Tuesday Afternoon." The booklet with vintage photos and insights adds a nice touch, making this a great package for both longtime fans and newcomers. Overall, it's a brilliant compilation that captures the magic of the Moody Blues perfectly. (0.861)
\\
\midrule
\midrule

\textbf{Review History}: \\
1 : Harry Nilsson has \textbf{an amazing voice} and his songs are very entertaining.  I love listening to him. \\
2 : This is \underline{a very good album}.  Both of them have \textbf{amazing voices} and they sound very good together. \\
3 : I never tire of listening to Dwight Yoakam and his music.  He has such \textbf{a great voice} for classic old country.  This is \underline{a fantastic collection} of his music. \\
4 : Great old classic stuff.  I never get tired of listening to these old bands.  Brings back a lot of memories \\
5 : Alan Jackson is one of my favorite country stars.  \textbf{His voice is really nice} and his songs are good. \\

\bottomrule
\end{tabular}
}
\end{table}

\subsection{RQ2: Number of Conversations and Negatives}

In the previous section, among the $n = 5$ user reviews, the most recent $\ell = n - 1$ reviews were converted into the conversational prompt as \texttt{assistant} responses, while the remaining one review was used in the first instruction text. In the \textbf{CCP} setting, a negative example is inserted at every turn.

This section investigates how the number of conversational turns $\ell$ and the number of inserted negative examples $m$ affect generation accuracy. We compare the following three settings: $(\ell, m) = (4, 4)$, which follows the same setup as in the previous section with a negative inserted at every turn; $(\ell, m) = (4, 1)$, in which $n - 1$ reviews are converted into conversational prompts but a negative is inserted only at the final turn; and $(\ell, m) = (1, 1)$, in which $n - 1$ reviews are included in the first instruction text and only one review is converted into a conversational prompt.

For the remaining experiments, to reduce inference costs under budget constraints, we limit the number of users per dataset from 200 to 50\footnote{Details of inference costs are discussed in the appendix.}. Instead, we merge all eight datasets and evaluate each method using a total of 400 users.

\begin{figure}[t]
    \centering
    \includegraphics[width=\linewidth]{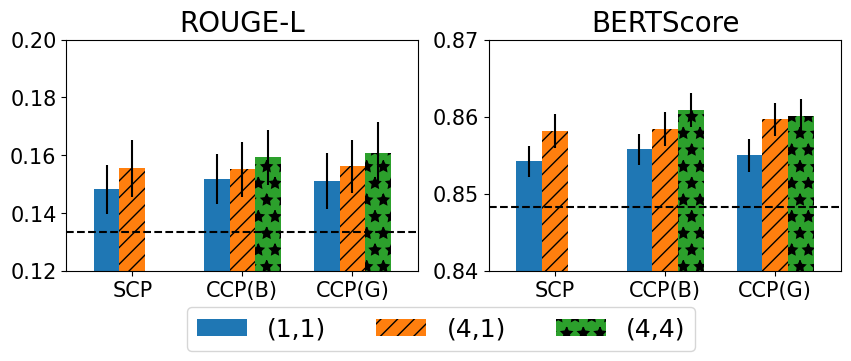}
    \caption{Scores for different conversations. Error bars indicate 95\% confidence intervals based on $t$-distribution. The dotted horizontal line indicates the score of the \textbf{Baseline}.}
    \label{fig:ablation_conversation}
\end{figure}

According to Figure~\ref{fig:ablation_conversation}, accuracies increased in the order of $(1,1) < (4,1) < (4,4)$, regardless of the prompt type or metric\footnote{Since \textbf{SCP} does not depend on the number of negatives $m$, the scores for $(4,4)$ and $(4,1)$ are the same, so $(4,4)$ is omitted.}. This indicates that converting each review into a conversational format and inserting a negative example at each turn is important for improving performance.

For \textbf{CCP}, collecting negative examples required additional cost, either from actual users or from an LLM. The setting $(4,1)$ introduced the smallest extra cost because it used only one negative example. However, the difference in accuracy compared with \textbf{SCP} at $\ell = 4$ was small. These results suggest that inserting negatives at every turn, as in $(\ell, m) = (n - 1, n - 1)$, is the most effective option if additional cost is acceptable. When the cost budget is limited, \textbf{SCP} with $\ell = n - 1$ offers a better trade-off between performance and cost.

\subsection{RQ3: Number of Reviews}

This study focuses on generating high-quality reviews for cold-start users who have only a few past reviews. While LLMs can capture a user's writing style when enough history is available~\citep{DBLP:conf/acl/role_llm_WangPQLZWGGN00024}, collecting many reviews per user is often difficult in real-world applications. Therefore, we examine how output quality changes as the number of past reviews varies by evaluating $n = 2, 3, 4, 5$, and $10$.

\begin{figure}[t]
    \centering
    \includegraphics[width=\linewidth]{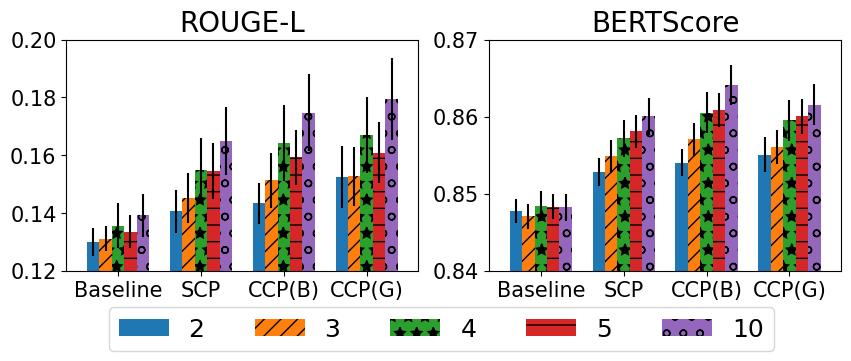}
    \caption{Scores for different review numbers. Error bars indicate 95\% confidence intervals based on $t$-distribution.}
    \label{fig:ablation_review}
\end{figure}

Figure~\ref{fig:ablation_review} showed that both \textbf{SCP} and \textbf{CCP} improved in accuracy as more reviews were used. This trend was observed in both evaluation metrics. In contrast, \textbf{Baseline} showed little improvement even as the number of reviews increased. These results suggest that the scaling laws of in-context learning~\citep{DBLP:conf/nips/incontext_AgarwalSZBRCZAA24}, which indicate that performance improves as more input examples are given, may not generalize well in a non-conversational prompt format.

In terms of absolute scores, both \textbf{SCP} and \textbf{CCP} achieved ROUGE scores over 0.14 and BERTScores over 0.85 with only $n=2$ reviews, which were higher than those of random user reviews. This suggests that even one conversational turn and a single negative example can improve the quality of the output. When $n=10$, ROUGE and BERTScore exceeded 0.16 and 0.86, showing that the generated reviews aligned more closely with the target user's style. These results suggest that simply increasing the number of reviews in a single prompt is not enough, and using a conversational prompt is important to take advantage of longer histories.

\subsection{RQ4: Effect on LLMs}

In previous sections, we used \texttt{gpt-4.1-mini} as the base LLM. This section examines whether the performance of different prompt types remains stable across various LLMs. We also evaluated \texttt{gpt-4.1} and \texttt{o4-mini} from OpenAI, \texttt{llama3.3-70b} from Meta, and \texttt{claude-sonnet-4} from Claude\footnote{The implementation details are summarized in the appendix, as in the case of \texttt{gpt-4.1-mini}.}. Among these, \texttt{o4-mini} is designed as a reasoning model that executes intermediate steps of reasoning before producing the final output.

\begin{figure}[t]
    \centering
    \includegraphics[width=\linewidth]{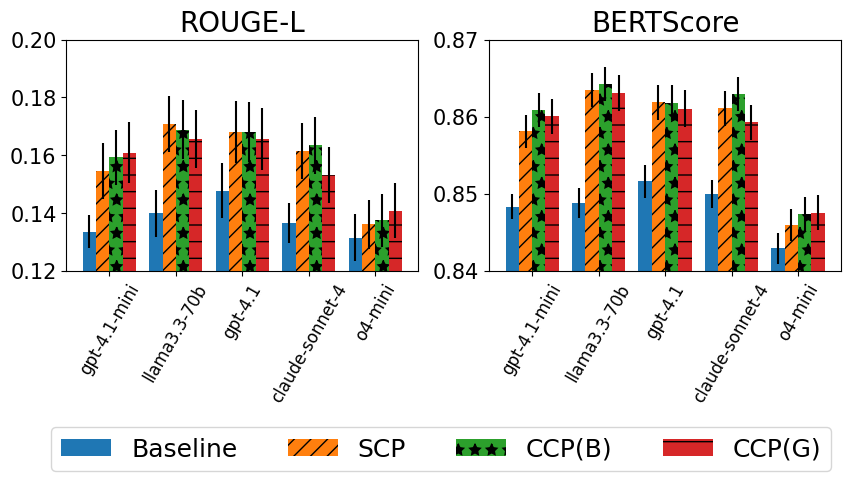}
    \caption{Scores for different LLMs. Error bars indicate 95\% confidence intervals based on $t$-distribution.}
    \label{fig:ablation_llms}
\end{figure}

We returned to the RQ2 setting with $n=5$ and 400 users. As shown in Figure~\ref{fig:ablation_llms}, both \textbf{SCP} and \textbf{CCP} consistently outperformed the \textbf{Baseline} across all LLMs. \texttt{o4-mini} had lower scores, possibly because its internal reasoning interfered with review generation, a phenomenon known as overthinking~\citep{DBLP:journals/corr/overthinking_abs-2503-16419,DBLP:journals/corr/overthinking_cot_abs-2410-21333}. Therefore, we excluded \texttt{o4-mini} from the subsequent analysis.

Among the \textbf{Baseline} results, \texttt{gpt-4.1} achieved the highest score, around 0.14 for ROUGE and 0.85 for BERTScore. In comparison, \textbf{SCP} achieved higher scores with all LLMs except \texttt{o4-mini}, which indicates that conversational prompting improved performance across different models. The inference costs of \texttt{gpt-4.1-mini} and \texttt{llama3.3-70b} were less than one-fourth of those for \texttt{gpt-4.1}. These results suggest that a cost-efficient LLM using conversational prompts can outperform non-conversational prompting with a more expensive model\footnote{Details of inference cost are provided in the appendix.}.

Among the \textbf{CCP} variants, \textbf{CCP(B)} outperformed \textbf{SCP} in many cases, while \textbf{CCP(G)} performed worse in some cases. However, the differences were small, and their confidence intervals overlapped. These observations suggest that \textbf{CCP(B)} is preferable when negative examples are available, whereas \textbf{SCP} is a better choice when they are not. Among the LLMs, \texttt{llama3.3-70b} generally achieved higher scores, but its confidence intervals also overlapped with those of the other LLMs. These results suggest that the effectiveness of conversational prompts remains stable across different LLMs.

\section{Task-Oriented Evaluation}

This study focuses on generating reviews with potential applications. This section examines the usefulness of the generated reviews for downstream tasks.

\subsection{User Identity Linkage}

To assess how well the generated reviews reflect the target user's writing style, we designed a matching task inspired by user identity linkage~\citep{DBLP:journals/sigkdd/uil_survey_ShuWTZL16, DBLP:conf/icwsm/uil_ZafaraniL09}, which aims to identify the author of a given text. For each item, we prepared three types of reviews: the ground-truth review, the generated review, and reviews from other users. We calculated the BERTScore between the ground-truth review and each of the others and ranked all reviews by the score to determine the position of the generated review.

As metrics, we used Hit@5, which measures how often the generated review appears in the top five, and mean reciprocal rank (MRR), which averages the reciprocal of the rank\footnote{Each item has at least five reviews from other users, with a median of 60 and an average of 285 reviews (Table~\ref{table:stat}), so we adopted Hit@5 as a metric. While Hit@5 is a binary indicator, MRR captures ranking accuracy on a continuous scale.}. To statistically evaluate these metrics, we computed confidence intervals using the bootstrap method with a resampling size of 1000. For comparison, we also included a baseline that randomly selects a review from other users.

\begin{figure}[t]
    \centering
    \includegraphics[width=\linewidth]{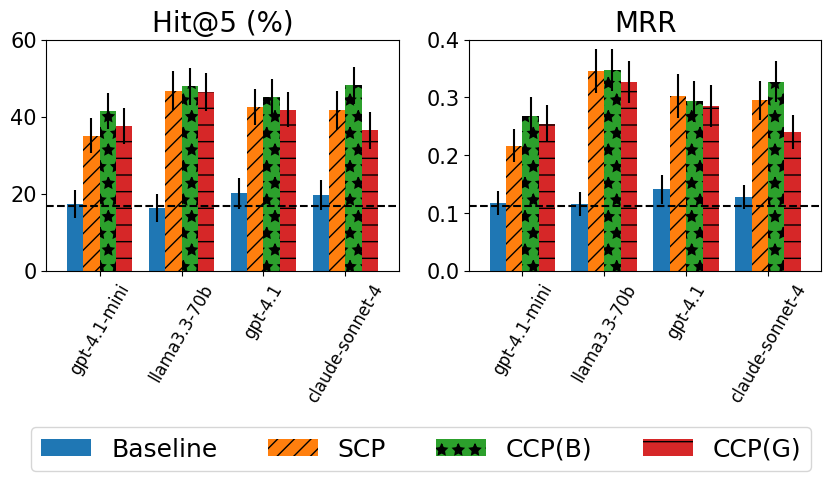}
    \caption{Hit@5 and MRR. Error bars indicate 95\% confidence intervals using the bootstrap method. The dotted horizontal line indicates the score of random reviews.}
    \label{fig:user_identity}
\end{figure}

As shown in Figure~\ref{fig:user_identity}, the \textbf{Baseline} performed about the same as random ranking in both Hit@5 and MRR across all LLMs. These results suggest that a non-conversational prompt did not capture the target user's writing style, even when using a strong model such as \texttt{gpt-4.1}. In contrast, conversational prompts improved both metrics. Among the conversational methods, \textbf{CCP(B)} achieved the highest scores, though the difference from \textbf{SCP} was small. This trend was consistent with the results shown in Figure~\ref{fig:ablation_llms}.

\subsection{Sentiment Analysis}

Product reviews include both positive and negative opinions, and each provides valuable insights for improving product development and marketing strategies. This section uses a RoBERTa-based sentiment classifier~\citep{barbieri-etal-2020-tweeteval}\footnote{\url{https://huggingface.co/cardiffnlp/twitter-roberta-base-sentiment-latest}} to assess how well the generated reviews reflect the sentiment tendencies of the target users. The classifier takes a text as input and returns a three-dimensional vector with scores for the sentiment labels: positive, neutral, and negative.

\subsubsection{Group-Level Evaluation}

\begin{figure}[t]
    \centering
    \includegraphics[width=\linewidth]{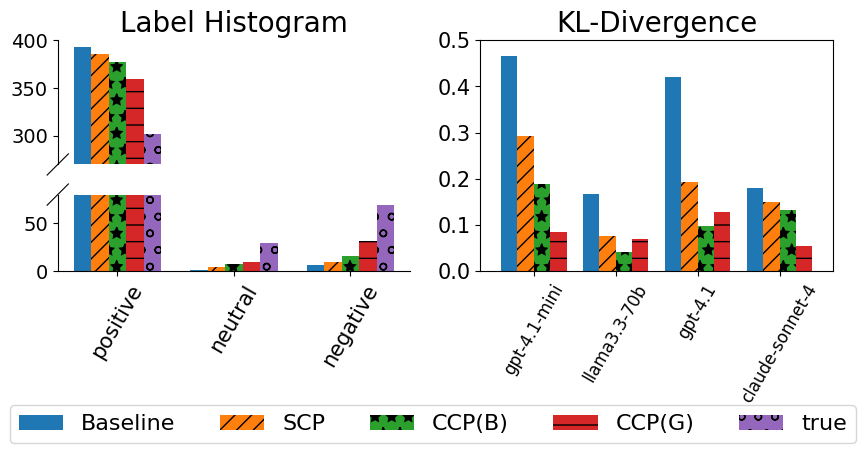}
    \caption{Sentiment label histogram by \texttt{gpt-4.1-mini} and KL divergence.}
    \label{fig:sentiment_group}
\end{figure}

Understanding the proportion of users with positive and negative opinions is an important factor in product development. This section examines how well the generated reviews reproduce that proportion. Specifically, for each user, we collect the sentiment label with the highest score, construct a histogram from these labels\footnote{The Amazon Reviews dataset includes a five-point rating scale (with 5 as most positive and 1 as most negative), whose distribution is 5: 259, 4: 67, 3: 35, 2: 16, 1: 23. Compared to Figure~\ref{fig:sentiment_group}, it can be seen that the classifier's sentiment labels generally reflect the original rating scores.}, and treat it as a distribution. We then evaluate how closely it matches the ground-truth histogram using the Kullback–Leibler (KL) divergence~\citep{DBLP:books/lib/Bishop06}.

As shown in the histogram in Figure~\ref{fig:sentiment_group}, reviews generated by the \textbf{Baseline} were strongly biased toward positive content, and few negative opinions were included, a pattern also discussed in \citep{DBLP:journals/corr/openreview_naacl_abs-2412-11948}. In contrast, conversational prompts generated more balanced reviews that included critical perspectives. The smaller KL divergence in the conversational prompts indicated that they reflected the user's sentiment tendencies more clearly.

\subsubsection{User-Level Evaluation}

We then evaluated how well the sentiment labels matched at the user level. As shown in Figure~\ref{fig:sentiment_group}, the dataset was highly imbalanced toward positive classes. Therefore, we used two metrics that are suitable for such situations: Weighted-F1, which averages F1 scores based on label frequency, and Macro-F1, which calculates the unweighted mean across all labels.

\begin{figure}[t]
    \centering
    \includegraphics[width=\linewidth]{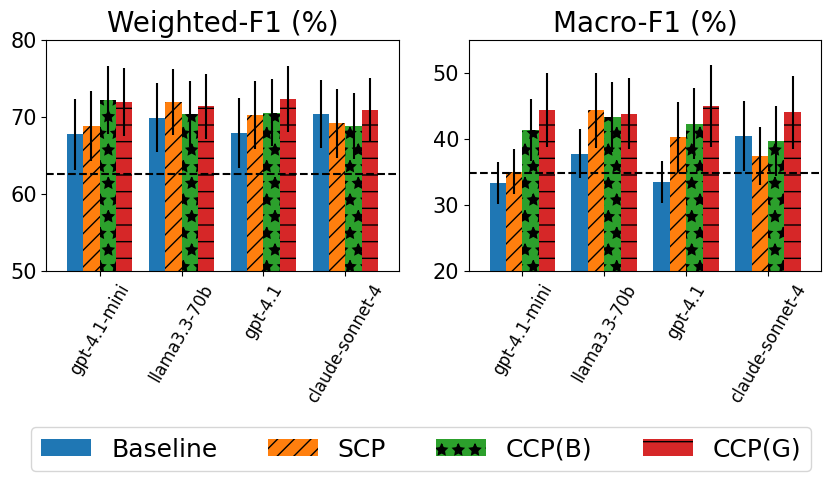}
    \caption{Weighted-F1 and Macro-F1. Error bars indicate 95\% confidence intervals using the bootstrap method. The dotted horizontal line indicates the score of random reviews.}
    \label{fig:sentiment_f1}
\end{figure}

As shown in Figure~\ref{fig:sentiment_f1}, conversational prompts achieved higher F1 scores than the \textbf{Baseline} for all models except \texttt{claude-sonnet-4}. This exception is probably because the baseline had a low KL divergence in Figure~\ref{fig:sentiment_group}, suggesting that the LLM could already reproduce the user's sentiment tendencies with reasonable accuracy. All prompts scored better than random reviews in Weighted-F1. On the other hand, the \textbf{Baseline} showed a lower score in Macro-F1. One possible reason is that the \textbf{Baseline} often failed to generate negative reviews, while conversational prompts produced a more balanced mix of positive and negative sentiments.

\section{Limitations and Discussion}

This study empirically showed that conversational prompting is effective for review generation. However, theoretical analysis is beyond the scope of this paper and is left for future work. Recent studies have examined the mechanisms of in-context learning with single prompts~\citep{DBLP:conf/iclr/icl_ChoKSI25,DBLP:journals/corr/icl_google,DBLP:conf/icml/icl_JeonLLR24}, and applying these insights to conversational formats could be promising.

One way to further improve accuracy is to modify \textbf{CCP(G)} so that it generates multiple reviews and selects the one with the highest similarity. Since \textbf{CCP(B)} outperformed \textbf{CCP(G)} in some cases, this method may improve results. However, generating multiple outputs increases inference cost, which can be several times higher than the cost of \textbf{SCP}. In situations where many users need to be processed without training, such as large-scale simulations~\citep{park2024generativeagentsimulations1000, DBLP:journals/corr/billion_agent_abs-2506-12078}, cost efficiency is also important. Exploring conversational prompts that reduce cost while maintaining or improving accuracy remains an area for future work.

Other possible applications include recommender systems~\citep{DBLP:conf/aaai/rag_review_Xie0MM23,DBLP:conf/emnlp/xrec_MaR024} and persona-based agents~\citep{DBLP:journals/corr/persona_survey_tmlr,DBLP:conf/acl/role_llm_WangPQLZWGGN00024}. These applications often require subjective human evaluation or integration into algorithms. We therefore leave their assessment for future work. Instead, this paper focuses on the internal quality of generated reviews and evaluates them using textual similarity, user identity linkage, and sentiment analysis across eight datasets and five LLMs.

\bibliography{aaai2026}

\section{Appendix}
\renewcommand{\thetable}{A\arabic{table}}
\setcounter{table}{0}
\renewcommand{\thefigure}{A\arabic{figure}} 
\setcounter{figure}{0} 

\subsection{Experimental setups}
Table~\ref{table:llms} summarizes the LLMs, including their versions and token prices. All OpenAI models were accessed through the official API (\url{https://platform.openai.com/docs/pricing}). Models from Meta and Anthropic were accessed via Amazon Bedrock (\url{https://aws.amazon.com/bedrock/pricing/}). All pricing in Table~\ref{table:llms} is as of July 30, 2025.

\begin{table}[t]
\centering
\caption{LLMs used in this study, including their version and token prices (USD per 1M tokens).}
\label{table:llms}
\scalebox{0.74}{
\begin{tabular}{lllcc}
\toprule
\textbf{Provider} & \textbf{Model} & \textbf{Version} & \textbf{Input} & \textbf{Output} \\
\midrule
OpenAI     & \texttt{gpt-4.1-mini}      & 2025-04-14       & 0.4  & 1.6  \\
OpenAI     & \texttt{gpt-4.1}           & 2025-04-14       & 2.0  & 8.0  \\
OpenAI     & \texttt{o4-mini}           & 2025-04-16       & 1.1  & 4.0  \\
Meta       & \texttt{llama3.3-70b}      & v1(2024-12-19)   & 0.72 & 0.72 \\
Anthropic  & \texttt{claude-sonnet-4}   & 20250514-v1      & 3.0  & 15   \\
\bottomrule
\end{tabular}
}
\end{table}

\subsection{Prompt}

Figure~\ref{list:ccp} shows an example of \textbf{CCP}. In this example, the first four items are used in the first instruction text, and the remaining one is converted into a conversational format with a rejection message. Using the notation of ``RQ2'' section, this corresponds to the case of $(\ell, m) = (4, 1)$. Figure~\ref{list:self-refine}
shows an example of \textbf{Self-Refine}. 

\begin{figure*}[t]
\centering
\begin{lstlisting}[frame=single]
{
    'role' : 'user', 'content' : """You are an AI assistant. As an AI assistant, I want you to impersonate me. I will provide you with product information, and I would like you to generate a review that I might write.
    
# My Information
Here is my previous item information. The items are listed in chronological order from 1 (oldest) to 4 (latest).
## Item history
{
    1: {
        'itemInfo': {'title': 'Angels And Alcohol', 'category': "Country, Today's Country", 'description': 'Alan Jackson\'s 15th studio album This year marks (*@\textcolor{blue}{(omitted)}@*)'},
        'review': 'Alan Jackson is one of my favorite country stars.  His voice is really nice and his songs are good.'
    }, 
    (*@\textcolor{blue}{(omitted)}@*), 
    4: {
        'itemInfo': {'title': 'Cheek to Cheek', 'category': 'Universal Music Group', 'description': "Deluxe edition includes three bonus tracks. (*@\textcolor{blue}{(omitted)}@*)"},
        'review': 'This is a very good album.  Both of them have amazing voices and they sound very good together.'
        }
}

# Task
Please predict the impressions I might have and generate a review for the target item.
## Target item
{'title': 'Son Of Schmilsson Expanded Edition', 'category': 'Pop, Singer-Songwriters', 'description': "Son followed in the footsteps of it's  (*@\textcolor{blue}{(omitted)}@*)"}

## Output format
Your response should only be the generated review. Do not include any irrelevant information."""  (*@\textcolor{red}{\# first instruction text}@*)
}
- - - - - - - - - - - - - - - - - - - - - - - - - - - - - - - - - - - - - - - - - - - -
{
    'role' : 'assistant', 'content' : """If you are a Nilsson fan and you don't own this album, I highly recommend it. I've loved it since (*@\textcolor{blue}{(omitted)}@*)"""  (*@\textcolor{red}{\# incorrect review}@*)
}
- - - - - - - - - - - - - - - - - - - - - - - - - - - - - - - - - - - - - - - - - - - -
{
    'role' : 'user', 'content' : """Absolutely different! That's not how I would answer. Please think it over carefully and generate a review for the target item that I might actually write."""  (*@\textcolor{red}{\# rejection message}@*)
}
- - - - - - - - - - - - - - - - - - - - - - - - - - - - - - - - - - - - - - - - - - - -
{
    'role' : 'assistant', 'content' : """Harry Nilsson has an amazing voice and his songs are very entertaining.  I love listening to him."""  (*@\textcolor{red}{\# true review}@*)
}
- - - - - - - - - - - - - - - - - - - - - - - - - - - - - - - - - - - - - - - - - - - -
{
    'role' : 'user', 'content' : """Excellent! It really feels like something I would write. Now, I will provide the next product. Please generate a review that I might write in the same way.
## Target item 
{'title': 'The Very Best Of The Moody Blues', 'category': 'Rock, Progressive, Progressive Rock', 'description': 'Product Description, This collection featuring hits from all phases of the Moody\'s career contains  (*@\textcolor{blue}{(omitted)}@*)'}"""  (*@\textcolor{red}{\# acceptance message}@*)
}
\end{lstlisting}
\caption{Detailed prompt for CCP}
\label{list:ccp}
\end{figure*}

\begin{figure*}[t]
\centering
\begin{lstlisting}[frame=single]
{
    'role' : 'user', 'content' : """You are an AI assistant. As an AI assistant, I want you to impersonate me. I will provide you with product information, and I would like you to generate a review that I might write.
    
# My Information
Here is my previous item information. The items are listed in chronological order from 1 (oldest) to 5 (latest).
## Item history
(*@\textcolor{blue}{(omitted)}@*)

# Task
Please predict the impressions I might have and generate a review for the target item.
## Target item
{'title': 'The Very Best Of The Moody Blues', 'category': 'Rock, Progressive, Progressive Rock', 'description': (*@\textcolor{blue}{(omitted)}@*)'}

## Output format
Your response should only be the generated review. Do not include any irrelevant information."""  (*@\textcolor{red}{\# first instruction text}@*)
}
- - - - - - - - - - - - - - - - - - - - - - - - - - - - - - - - - - - - - - - - - - - -
{
    'role' : 'assistant', 'content' : """The Moody Blues have such a unique sound that really stands out in rock music.  This collection is (*@\textcolor{blue}{(omitted)}@*)"""  (*@\textcolor{red}{\# generated review}@*)
}
- - - - - - - - - - - - - - - - - - - - - - - - - - - - - - - - - - - - - - - - - - - -
{
    'role' : 'user', 'content' : """Thank you! However, could you please critically review the created review from the perspective of whether it reflects the user's past review style and way of thinking, and suggest how it could be improved further?"""  
    (*@\textcolor{red}{\# refinement message}@*)
}
- - - - - - - - - - - - - - - - - - - - - - - - - - - - - - - - - - - - - - - - - - - -
{
    'role' : 'assistant', 'content' : """Certainly! Here's a critical review of the generated review based on your past review style and suggestions for improvement:
1. **Tone and Style Consistency**  
Your previous reviews tend to be straightforward, positive, and personal, often highlighting what you like about the artist or album in simple, clear language. For example, (*@\textcolor{blue}{(omitted)}@*)
2. **Level of Detail**  
(*@\textcolor{blue}{(omitted)}@*)
---
### Suggestions for Improvement
- Add a more personal touch by mentioning what you enjoy about The Moody Blues specifically (e.g., their melodies, songwriting, or how their music makes you feel).  
- (*@\textcolor{blue}{(omitted)}@*)
---
### Improved Version Example
The Moody Blues have such a distinctive sound that I really enjoy. (*@\textcolor{blue}{(omitted)}@*)
---
This version better reflects your past review style by being personal, positive, and simple while still mentioning key features of the album."""  (*@\textcolor{red}{\# thought}@*)
}
- - - - - - - - - - - - - - - - - - - - - - - - - - - - - - - - - - - - - - - - - - - -
{
    'role' : 'user', 'content' : """'OK. Then, taking your critique into account, please rewrite the review. Remember, your response should be only the generated review; do not include any irrelevant information."""  (*@\textcolor{red}{\# request message}@*)
}
\end{lstlisting}
\caption{Detailed prompt for Self-Refine}
\label{list:self-refine}
\end{figure*}

\subsection{RQ1}

Table~\ref{table:rq1_full} adds the results of \textbf{Self-Refine}, which were not included in Table 3 in the main paper. The score for \textbf{SCP+SR} refers to the output generated by applying \textbf{Self-Refine} to the review created by \textbf{SCP}, following the procedure shown in Figure~\ref{list:self-refine}. As noted in the footnote of the main paper, applying \textbf{Self-Refine} to reviews generated by \textbf{SCP} or \textbf{CCP(G)} did not lead to better accuracy. As indicated by the $\diamond$ marks, performance often dropped to the level of the \textbf{Baseline}.

\begin{table*}[t]
\caption{ROUGE-L and BERTScore, where bold and underlined values represent best and second-best methods in each column. The column names ``R'' and ``B'' refer to ROUGE-L and BERTScore, respectively. Prompts significantly better than \textbf{SCP} are marked with $*$, while those not significantly better than \textbf{Baseline} are marked with $\diamond$ (one-sided Wilcoxon test, $p < 0.01$).}
\label{table:rq1_full}
\centering
\scalebox{0.65}{
\begin{tabular}{c|cc|cc|cc|cc|cc|cc|cc|cc||cc}
\toprule
 & \multicolumn{2}{|c}{Movies} & \multicolumn{2}{c}{Music} & \multicolumn{2}{c}{Books} & \multicolumn{2}{c}{Kindle} & \multicolumn{2}{c}{Groceries} & \multicolumn{2}{c}{Games} & \multicolumn{2}{c}{Sports} & \multicolumn{2}{c}{Electronics} & \multicolumn{2}{||c}{All} \\ \midrule
 & R & B & R & B & R & B & R & B & R & B & R & B & R & B & R & B & R & B\\
\midrule
Baseline & $0.126$ & $0.844$ & $0.129$ & $0.847$ & $0.148$ & $0.852$ & $0.137$ & $0.852$ & $0.132$ & $0.846$ & $0.137$ & $0.850$ & $0.137$ & $0.848$ & $0.123$ & $0.843$ & $0.133$ & $0.848$ \\
Self-Refine & $0.141$ & $0.845^{\diamond}$ & $0.144$ & $0.849^{\diamond}$ & $0.157^{\diamond}$ & $0.854^{\diamond}$ & $0.134^{\diamond}$ & $0.850^{\diamond}$ & $0.127^{\diamond}$ & $0.845^{\diamond}$ & $0.137^{\diamond}$ & $0.851^{\diamond}$ & $0.130^{\diamond}$ & $0.846^{\diamond}$ & $0.123^{\diamond}$ & $0.842^{\diamond}$ & $0.137^{\diamond}$ & $0.848^{\diamond}$ \\ \midrule
SCP & $0.153$ & $0.854$ & $0.165$ & $0.859$ & $0.183$ & $0.864$ & $0.154$ & $0.860$ & $0.143$ & $0.854$ & $0.152$ & $0.860$ & $0.149$ & $0.857$ & $0.134$ & $0.850$ & $0.154$ & $0.857$ \\
CCP(B) & $\underline{0.171}^{*}$ & $\textbf{0.860}^{*}$ & $\underline{0.177}$ & $\underline{0.864}^{*}$ & $0.186$ & $\underline{0.866}^{*}$ & $\underline{0.159}$ & $\textbf{0.863}^{*}$ & $0.148$ & $\textbf{0.857}^{*}$ & $\textbf{0.157}$ & $\textbf{0.863}^{*}$ & $\underline{0.152}$ & $\textbf{0.858}^{*}$ & $0.136$ & $\textbf{0.853}^{*}$ & $\underline{0.161}^{*}$ & $\textbf{0.861}^{*}$ \\
CCP(R) & $0.167^{*}$ & $0.857^{*}$ & $0.175$ & $0.863^{*}$ & $\underline{0.187}$ & $0.866^{*}$ & $\textbf{0.161}$ & $\underline{0.863}^{*}$ & $\underline{0.149}$ & $\underline{0.857}^{*}$ & $\underline{0.155}$ & $0.861^{*}$ & $0.150$ & $\underline{0.857}$ & $\textbf{0.138}$ & $\underline{0.852}^{*}$ & $0.160^{*}$ & $0.860^{*}$ \\
CCP(G) & $\textbf{0.175}^{*}$ & $\underline{0.858}^{*}$ & $\textbf{0.187}^{*}$ & $\textbf{0.865}^{*}$ & $\textbf{0.191}^{*}$ & $\textbf{0.867}^{*}$ & $0.159$ & $0.863^{*}$ & $\textbf{0.150}$ & $0.856$ & $0.155$ & $\underline{0.861}$ & $\textbf{0.158}$ & $0.857$ & $\underline{0.137}$ & $0.850$ & $\textbf{0.164}^{*}$ & $\underline{0.860}^{*}$ \\
CCP(B)$^{-}$ & $0.152$ & $0.854$ & $0.172$ & $0.862$ & $0.179$ & $0.863$ & $0.155$ & $0.861$ & $0.141$ & $0.854$ & $0.153$ & $0.861$ & $0.149$ & $0.857$ & $0.135$ & $0.851^{*}$ & $0.154$ & $0.858^{*}$ \\
CCP(R)$^{-}$ & $0.163$ & $0.857^{*}$ & $0.173$ & $0.862$ & $0.186$ & $0.866^{*}$ & $0.155$ & $0.862^{*}$ & $0.145$ & $0.856^{*}$ & $0.154$ & $0.861$ & $0.150$ & $0.857$ & $0.133$ & $0.851^{*}$ & $0.157$ & $0.859^{*}$ \\ \midrule
SCP+SR & $0.140$ & $0.846$ & $0.145$ & $0.850$ & $0.149^{\diamond}$ & $0.854$ & $0.140^{\diamond}$ & $0.853^{\diamond}$ & $0.133^{\diamond}$ & $0.848$ & $0.143$ & $0.853$ & $0.131^{\diamond}$ & $0.848^{\diamond}$ & $0.126^{\diamond}$ & $0.845$ & $0.138$ & $0.850$ \\
CCP(B)+SR & $0.142$ & $0.848$ & $0.139$ & $0.849$ & $0.151^{\diamond}$ & $0.855$ & $0.142$ & $0.853$ & $0.129^{\diamond}$ & $0.847$ & $0.143$ & $0.854$ & $0.132^{\diamond}$ & $0.849$ & $0.129$ & $0.845$ & $0.138$ & $0.850$ \\
CCP(R)+SR & $0.143$ & $0.847$ & $0.137^{\diamond}$ & $0.849^{\diamond}$ & $0.149^{\diamond}$ & $0.854$ & $0.138^{\diamond}$ & $0.853$ & $0.134^{\diamond}$ & $0.848$ & $0.140^{\diamond}$ & $0.853$ & $0.135^{\diamond}$ & $0.849^{\diamond}$ & $0.128^{\diamond}$ & $0.845$ & $0.138$ & $0.850$ \\
CCP(G)+SR & $0.144$ & $0.847$ & $0.143$ & $0.849^{\diamond}$ & $0.150^{\diamond}$ & $0.854$ & $0.139^{\diamond}$ & $0.853$ & $0.132^{\diamond}$ & $0.847^{\diamond}$ & $0.141^{\diamond}$ & $0.853$ & $0.134^{\diamond}$ & $0.848^{\diamond}$ & $0.126^{\diamond}$ & $0.844$ & $0.138$ & $0.849$ \\
CCP(B)$^{-}$+SR & $0.142$ & $0.847$ & $0.142$ & $0.849$ & $0.150^{\diamond}$ & $0.854$ & $0.137^{\diamond}$ & $0.852^{\diamond}$ & $0.130^{\diamond}$ & $0.846^{\diamond}$ & $0.145$ & $0.854$ & $0.138^{\diamond}$ & $0.849^{\diamond}$ & $0.129$ & $0.845$ & $0.139$ & $0.849$ \\
CCP(R)$^{-}$+SR & $0.141$ & $0.847$ & $0.137$ & $0.848^{\diamond}$ & $0.148^{\diamond}$ & $0.854$ & $0.141^{\diamond}$ & $0.854$ & $0.130^{\diamond}$ & $0.848$ & $0.141^{\diamond}$ & $0.854$ & $0.134^{\diamond}$ & $0.848^{\diamond}$ & $0.128$ & $0.845$ & $0.138$ & $0.850$ \\
\bottomrule
\end{tabular}
}
\end{table*}

Table~\ref{table:rq1_cost} shows the inference cost for each method using \texttt{gpt-4.1-mini}. For entries with \textbf{SR}[$\cdot$], the cost reflects only the \textbf{Self-Refine} step and does not include the cost of generating the initial review with methods such as \textbf{SCP}. For example, across all datasets, \textbf{CCP(G)} incurred a cost of \$5.50 to generate reviews, and \textbf{Self-Refine} required an additional \$1.69 to revise them. Therefore, obtaining reviews using \textbf{CCP(G)+SR} required a total cost of \$5.50 + \$1.69 = \$7.29.

\begin{table*}[t]
\caption{API fee (US dollars)}
\label{table:rq1_cost}
\centering
\scalebox{1}{
\begin{tabular}{c|cccccccc|cc}
\toprule
 & Movies & Music & Books & Kindle & Groceries & Games & Sports & Electronics & All \\ 
\midrule
Baseline & $0.15$ & $0.14$ & $0.17$ & $0.17$ & $0.14$ & $0.16$ & $0.16$ & $0.18$ & $1.27$ \\
SCP & $0.15$ & $0.14$ & $0.16$ & $0.17$ & $0.14$ & $0.16$ & $0.16$ & $0.17$ & $1.25$ \\
CCP(B) & $0.17$ & $0.16$ & $0.19$ & $0.19$ & $0.16$ & $0.18$ & $0.18$ & $0.19$ & $1.42$ \\
CCP(R) & $0.18$ & $0.17$ & $0.19$ & $0.20$ & $0.17$ & $0.18$ & $0.19$ & $0.20$ & $1.46$ \\
CCP(G) & $0.66$ & $0.64$ & $0.71$ & $0.72$ & $0.63$ & $0.70$ & $0.70$ & $0.74$ & $5.50$ \\
CCP(B)$^{-}$ & $0.19$ & $0.18$ & $0.21$ & $0.22$ & $0.17$ & $0.19$ & $0.19$ & $0.21$ & $1.56$ \\
CCP(R)$^{-}$ & $0.17$ & $0.16$ & $0.19$ & $0.19$ & $0.16$ & $0.18$ & $0.18$ & $0.19$ & $1.43$ \\ \midrule
SR[Baseline] & $0.21$ & $0.20$ & $0.23$ & $0.23$ & $0.21$ & $0.22$ & $0.23$ & $0.24$ & $1.77$ \\
SR[SCP] & $0.20$ & $0.19$ & $0.21$ & $0.22$ & $0.19$ & $0.21$ & $0.21$ & $0.23$ & $1.67$ \\
SR[CCP(B)] & $0.20$ & $0.19$ & $0.21$ & $0.22$ & $0.19$ & $0.21$ & $0.21$ & $0.23$ & $1.65$ \\
SR[CCP(R)] & $0.20$ & $0.19$ & $0.21$ & $0.22$ & $0.19$ & $0.21$ & $0.21$ & $0.23$ & $1.66$ \\
SR[CCP(G)] & $0.20$ & $0.19$ & $0.22$ & $0.22$ & $0.20$ & $0.21$ & $0.22$ & $0.23$ & $1.69$ \\
SR[CCP(B)$^{-}$] & $0.20$ & $0.19$ & $0.22$ & $0.22$ & $0.19$ & $0.21$ & $0.21$ & $0.23$ & $1.67$ \\
SR[CCP(R)$^{-}$] & $0.20$ & $0.19$ & $0.21$ & $0.22$ & $0.19$ & $0.21$ & $0.21$ & $0.23$ & $1.66$ \\
\bottomrule
\end{tabular}
}
\end{table*}

Tables \ref{table:rq1_over09} and \ref{table:rq1_over10} show the proportion of users (200 per dataset) for whom a review with BERTScore above 0.9 or exactly 1 was generated.

\begin{table}[t]
\caption{The number of generated reviews that BERTScore $>0.9$.}
\label{table:rq1_over09}
\centering
\scalebox{0.9}{
\begin{tabular}{c|cccc}
\toprule
 & Baseline & SCP & CCP(B) & CCP(G) \\
\midrule
Movies & 2 & 4 & 7 & 10 \\
Music & 3 & 7 & 14 & 16 \\
Books & 3 & 8 & 11 & 11 \\
Kindle & 2 & 3 & 5 & 3 \\
Groceries & 0 & 0 & 1 & 1 \\
Games & 0 & 1 & 1 & 2 \\
Sports & 1 & 2 & 2 & 3 \\
Electronics & 0 & 0 & 0 & 0 \\
\bottomrule
\end{tabular}
}
\end{table}

\begin{table}[t]
\caption{The number of exact matched generated reviews}
\label{table:rq1_over10}
\centering
\scalebox{0.9}{
\begin{tabular}{c|cccc}
\toprule
 & Baseline & SCP & CCP(B) & CCP(G) \\
\midrule
Movies & 0 & 1 & 3 & 4 \\
Music & 0 & 3 & 7 & 6 \\
Books & 0 & 6 & 6 & 5 \\
Kindle & 0 & 1 & 1 & 1 \\
Groceries & 0 & 0 & 0 & 0 \\
Games & 0 & 0 & 0 & 0 \\
Sports & 1 & 2 & 2 & 3 \\
Electronics & 0 & 0 & 0 & 0 \\
\bottomrule
\end{tabular}
}
\end{table}

\subsection{RQ2}

Tables~\ref{table:rq2_rouge} and \ref{table:rq2_bert} provide the numerical data that form Figure~3 in the main paper.

\begin{table}[t]
\caption{ROUGE-L (RQ2)}
\label{table:rq2_rouge}
\centering
\scalebox{1}{
\begin{tabular}{c|ccc}
\toprule
 & SCP & CCP(B) & CCP(G) \\
\midrule
(1,1) & $0.148$ & $0.152$ & $0.151$ \\
(4,1) & $0.155$ & $0.155$ & $0.156$ \\
(4,4) & - & $\underline{0.159}$ & $\textbf{0.161}$ \\
\bottomrule
\end{tabular}
}
\end{table}

\begin{table}[t]
\caption{BERTScore (RQ2)}
\label{table:rq2_bert}
\centering
\scalebox{1}{
\begin{tabular}{c|ccc}
\toprule
 & SCP & CCP(B) & CCP(G) \\
\midrule
(1,1) & $0.854$ & $0.856$ & $0.855$ \\
(4,1) & $0.858$ & $0.858$ & $0.860$ \\
(4,4) & - & $\textbf{0.861}$ & $\underline{0.860}$ \\
\bottomrule
\end{tabular}
}
\end{table}

\subsection{RQ3}

Tables~\ref{table:rq3_rouge} and \ref{table:rq3_bert} provide the numerical data that form Figure~4 in the main paper.

\begin{table}[t]
\caption{ROUGE-L (RQ3)}
\label{table:rq3_rouge}
\centering
\scalebox{0.9}{
\begin{tabular}{c|cccc}
\toprule
 & Baseline & SCP & CCP(B) & CCP(G) \\
\midrule
2 & $0.130$ & $0.141$ & $\underline{0.143}$ & $\textbf{0.152}$ \\
3 & $0.131$ & $0.145$ & $\underline{0.151}$ & $\textbf{0.153}$ \\
4 & $0.136$ & $0.155$ & $\underline{0.164}$ & $\textbf{0.167}$ \\
5 & $0.133$ & $0.155$ & $\underline{0.159}$ & $\textbf{0.161}$ \\
10 & $0.139$ & $0.165$ & $\underline{0.175}$ & $\textbf{0.179}$ \\
\bottomrule
\end{tabular}
}
\end{table}

\begin{table}[t]
\caption{BERTScore (RQ3)}
\label{table:rq3_bert}
\centering
\scalebox{0.9}{
\begin{tabular}{c|cccc}
\toprule
 & Baseline & SCP & CCP(B) & CCP(G) \\
\midrule
2 & $0.848$ & $0.853$ & $\underline{0.854}$ & $\textbf{0.855}$ \\
3 & $0.847$ & $0.855$ & $\textbf{0.857}$ & $\underline{0.856}$ \\
4 & $0.848$ & $0.857$ & $\textbf{0.861}$ & $\underline{0.859}$ \\
5 & $0.848$ & $0.858$ & $\textbf{0.861}$ & $\underline{0.860}$ \\
10 & $0.848$ & $0.860$ & $\textbf{0.864}$ & $\underline{0.862}$ \\
\bottomrule
\end{tabular}
}
\end{table}

\subsection{RQ4}

Tables~\ref{table:rq4_rouge} and \ref{table:rq4_bert} provide the numerical data that form Figure~5 in the main paper.

\begin{table}[t]
\caption{ROUGE-L (RQ4)}
\label{table:rq4_rouge}
\centering
\scalebox{0.9}{
\begin{tabular}{c|cccc}
\toprule
 & Baseline & SCP & CCP(B) & CCP(G) \\
\midrule
gpt-4.1-mini & $0.133$ & $0.155$ & $0.159$ & $0.161$ \\
llama3.3-70b & $0.140$ & $\textbf{0.171}$ & $\underline{0.169}$ & $0.166$ \\
gpt-4.1 & $0.148$ & $0.168$ & $0.168$ & $0.165$ \\
claude-sonnet-4 & $0.136$ & $0.161$ & $0.163$ & $0.153$ \\
o4-mini & $0.131$ & $0.136$ & $0.137$ & $0.141$ \\
\bottomrule
\end{tabular}
}
\end{table}

\begin{table}[t]
\caption{BERTScore (RQ4)}
\label{table:rq4_bert}
\centering
\scalebox{0.9}{
\begin{tabular}{c|cccc}
\toprule
 & Baseline & SCP & CCP(B) & CCP(G) \\
\midrule
gpt-4.1-mini & $0.848$ & $0.858$ & $0.861$ & $0.860$ \\
llama3.3-70b & $0.849$ & $\underline{0.863}$ & $\textbf{0.864}$ & $0.863$ \\
gpt-4.1 & $0.852$ & $0.862$ & $0.862$ & $0.861$ \\
claude-sonnet-4 & $0.850$ & $0.861$ & $0.863$ & $0.859$ \\
o4-mini & $0.843$ & $0.846$ & $0.847$ & $0.848$ \\
\bottomrule
\end{tabular}
}
\end{table}

Table~\ref{table:rq4_fee} shows the inference cost for the RQ4 experiments in the main paper. The values indicate the total cost for 400 users, with 50 sampled from each of the eight datasets. The combined cost across all LLMs was \$56.05. Running the same setting for 1,600 users would exceed \$200.

\begin{table}[t]
\caption{API fee (US dollars)}
\label{table:rq4_fee}
\centering
\scalebox{0.8}{
\begin{tabular}{c|cccc|c}
\toprule
 & Baseline & SCP & CCP(B) & CCP(G) & All \\
\midrule
gpt-4.1-mini & $0.32$ & $0.31$ & $0.35$ & $1.37$ & $2.36$ \\
llama3.3-70b & $0.49$ & $0.51$ & $0.59$ & $2.14$ & $3.73$ \\
gpt-4.1 & $1.64$ & $1.54$ & $1.78$ & $6.91$ & $11.87$ \\
claude-sonnet-4 & $3.00$ & $2.80$ & $3.09$ & $12.57$ & $21.47$ \\
o4-mini & $2.37$ & $1.87$ & $2.25$ & $10.13$ & $16.63$ \\
\bottomrule
\end{tabular}
}
\end{table}

\subsection{User Identity Linkage}

Tables~\ref{table:51_hit} and \ref{table:51_MRR} provide the numerical data that form Figure~6 in the main paper.

\begin{table}[t]
\caption{Hit@5 (\%)}
\label{table:51_hit}
\centering
\scalebox{0.9}{
\begin{tabular}{c|cccc}
\toprule
 & Baseline & SCP & CCP(B) & CCP(G) \\
\midrule
gpt-4.1-mini & $17.2$ & $35.0$ & $41.5$ & $37.5$ \\
llama3.3-70b & $16.2$ & $46.8$ & $\underline{47.8}$ & $46.5$ \\
gpt-4.1 & $20.0$ & $42.5$ & $45.0$ & $41.8$ \\
claude-sonnet-4 & $19.8$ & $41.5$ & $\textbf{48.2}$ & $36.5$ \\ \midrule
random & \multicolumn{4}{|c}{$16.7$} \\
\bottomrule
\end{tabular}
}
\end{table}

\begin{table}[t]
\caption{MRR}
\label{table:51_MRR}
\centering
\scalebox{0.9}{
\begin{tabular}{c|cccc}
\toprule
 & Baseline & SCP & CCP(B) & CCP(G) \\
\midrule
gpt-4.1-mini & $0.117$ & $0.216$ & $0.268$ & $0.254$ \\
llama3.3-70b & $0.116$ & $\underline{0.346}$ & $\textbf{0.347}$ & $0.326$ \\
gpt-4.1 & $0.140$ & $0.303$ & $0.293$ & $0.285$ \\
claude-sonnet-4 & $0.127$ & $0.295$ & $0.328$ & $0.240$ \\ \midrule
random & \multicolumn{4}{|c}{$0.112$} \\
\bottomrule
\end{tabular}
}
\end{table}

\subsection{Sentiment Analysis}

Tables~\ref{table:52_histogram} and \ref{table:52_KL} provide the numerical data that form Figure~7 in the main paper.

\begin{table}[t]
\caption{Sentiment label histogram by \texttt{gpt-4.1-mini}}
\label{table:52_histogram}
\centering
\scalebox{0.9}{
\begin{tabular}{c|cccc|c}
\toprule
 & Baseline & SCP & CCP(B) & CCP(G) & True \\
\midrule
positive & $393$ & $386$ & $377$ & $359$ & $302$ \\
neutral & $1$ & $4$ & $7$ & $10$ & $29$ \\
negative & $6$ & $10$ & $16$ & $31$ & $69$ \\
\bottomrule
\end{tabular}
}
\end{table}

\begin{table}[t]
\caption{KL divergence}
\label{table:52_KL}
\centering
\scalebox{0.9}{
\begin{tabular}{c|cccc}
\toprule
 & Baseline & SCP & CCP(B) & CCP(G) \\
\midrule
gpt-4.1-mini & $0.467$ & $0.292$ & $0.188$ & $0.085$ \\
llama3.3-70b & $0.166$ & $0.076$ & $0.042$ & $0.069$ \\
gpt-4.1 & $0.420$ & $0.193$ & $0.098$ & $0.128$ \\
claude-sonnet-4 & $0.180$ & $0.150$ & $0.132$ & $0.055$ \\
\bottomrule
\end{tabular}
}
\end{table}

Tables~\ref{table:52_Weighted} and \ref{table:52_macro} provide the numerical data that form Figure~8 in the main paper.

\begin{table}[t]
\caption{Weighted-F1 (\%)}
\label{table:52_Weighted}
\centering
\scalebox{0.9}{
\begin{tabular}{c|cccc}
\toprule
 & Baseline & SCP & CCP(B) & CCP(G) \\
\midrule
gpt-4.1-mini & $67.7$ & $68.7$ & $\underline{72.2}$ & $71.9$ \\
llama3.3-70b & $69.9$ & $71.9$ & $70.4$ & $71.3$ \\
gpt-4.1 & $67.8$ & $70.2$ & $70.5$ & $\textbf{72.3}$ \\
claude-sonnet-4 & $70.3$ & $69.1$ & $68.8$ & $70.9$ \\ \midrule
random & \multicolumn{4}{|c}{$62.5$} \\
\bottomrule
\end{tabular}
}
\end{table}

\begin{table}[t]
\caption{Macro-F1 (\%)}
\label{table:52_macro}
\centering
\scalebox{0.9}{
\begin{tabular}{c|cccc}
\toprule
 & Baseline & SCP & CCP(B) & CCP(G) \\
\midrule
gpt-4.1-mini & $33.3$ & $34.9$ & $41.2$ & $\underline{44.4}$ \\
llama3.3-70b & $37.7$ & $44.3$ & $43.3$ & $43.8$ \\
gpt-4.1 & $33.4$ & $40.2$ & $42.2$ & $\textbf{44.9}$ \\
claude-sonnet-4 & $40.3$ & $37.4$ & $39.6$ & $43.9$ \\ \midrule
random & \multicolumn{4}{|c}{$34.7$} \\
\bottomrule
\end{tabular}
}
\end{table}

\end{document}